# *Top-k* Multi-Armed Bandit Learning for Content Dissemination in Swarms of Micro-UAVs

Amit Kumar Bhuyan, Hrishikesh Dutta, and Subir Biswas

***Abstract* — This paper presents a Micro-Unmanned Aerial Vehicle (UAV)-enhanced content management system for disaster scenarios where communication infrastructure is generally compromised. Utilizing a hybrid network of stationary and mobile Micro-UAVs, this system aims to provide crucial content access to isolated communities. In the developed architecture, stationary anchor UAVs, equipped with vertical and lateral links, serve users in individual disaster-affected communities. and mobile micro-ferrying UAVs, with enhanced mobility, extend coverage across multiple such communities. The primary goal is to devise a content dissemination system that dynamically learns caching policies to maximize content accessibility to users left without communication infrastructure. The core contribution is an adaptive content dissemination framework that employs a decentralized *Top-k* Multi-Armed Bandit learning approach for efficient UAV caching decisions. This approach accounts for geo-temporal variations in content popularity and diverse user demands. Additionally, a Selective Caching Algorithm is proposed to minimize redundant content copies by leveraging inter-UAV information sharing. Through functional verification and performance evaluation, the proposed framework demonstrates improved system performance and adaptability across varying network sizes, micro-UAV swarms, and content popularity distributions.**

## I. INTRODUCTION

Disasters such as earthquakes, floods, wars, and other catastrophic events can have devastating effects on people's lives and properties, as well as communication infrastructures. In such situations, people may be forced to migrate to areas without proper communication infrastructure, leaving them without access to important information such as the state of the disaster, rescue and relief operations, weather reports, rehabilitation efforts, etc. Especially in cluttered communities of users, such as migrating population or restoration camps that create hot spots for content demand, a large number of users need to download data from the ground base stations (BS). In this case, ensuring the quality of service (QoS) through content delivery is particularly important. Due to the nonstationary nature of the migrating population and varying user preferences, the user demands for content might be different. To address the content delivery problem in such scenarios, extensive research has been conducted by proposing device-to-device (D2D) communication [15], roadside units (RSU) [22] and small base stations (SBS) [17]. Given the failure of backhaul connectivity in such scenarios, the effectiveness of D2D communication is constrained by the total content diversity across all users. It is difficult to guarantee or even quantify the content diversity in scenarios like disaster, which does not generally entail apriori preparedness. Furthermore, while unavailability of backhaul connectivity also holds for RSUs and SBSs, their applicability is further hindered by their installation rigor and restricted mobility.

The authors are with Electrical and Computer Engineering Department of Michigan State University, East Lansing, MI, USA, 48823. (Email: *bhuyanam@msu.edu, duttahr1@msu.edu, sbiswas@msu.edu*)

This paper proposes the use of Micro-Unmanned Aerial Vehicles (Micro-UAVs) as an alternative content provisioning platform when fixed communication infrastructure such as cellular phone towers is unavailable. Their unrestricted mobility makes them better suited for the problem in hand. They are capable of taking the role of a mobile SBS or RSU, while their unrestricted mobility mitigates the limitations of SBS or RSU. Additionally, their small form factor allows them to maintain a low altitude flight pattern, therefore mitigating the need for long range communication. Avoiding such communication equipment has multifaceted benefits, namely low power consumption and relatively lighter payload. Micro-UAVs, however, bring their own limitations in terms of storage capacity, flight time, etc., which add new challenges to UAV based content storage and dissemination system.

This study introduces swarm of Micro-UAVs as content carriers in a dissemination system that uses a variant of Multi-armed Bandit Learning to perform optimal caching in communication-challenged environments. With the rapid surge in miniaturized UAV technology, their viability and usage has increased because of their low cost and [1], [2] small footprints. Studies have also shown that due to their relatively smaller size, the power dissipation and recharge time of micro-UAVs are lower as compared to larger UAVs [1]. Additionally, their low altitude flight capability makes them apt for unimpeded short-range communication which can be a key in post-disaster scenarios. These attributes have motivated this work, where a UAV-aided content dissemination system uses a large population of low-cost micro-UAVs for enhanced content availability in the absence of fixed communication infrastructure.

The framework discussed herewithin specifically focuses on scenarios where disaster/war-stricken populations are stranded and geographically clustered into multiple communities that may not have access to surviving cellular base stations. In such scenarios, the request patterns at different communities and the tolerable access delay (*TAD*) [4] can be different for different contents based on the type and urgency of the requested information. The proposed MAB learning solution deploys UAV/micro-UAV-based tactical content service provisioning that can make caching decisions on the fly without prior knowledge of content request pattern.

The proposed content provisioning system uses a two-tier architecture consisting of relatively larger anchor-UAVs (A-UAVs) and Micro-ferrying-UAVs (MF-UAVs). Each disaster-isolated user community is served by a A-UAV, which possesses expensive vertical connectivity, such as a satellite link [5], to download content. These A-UAVs also have lateral links to communicate with MF-UAVs and to disseminate contents to the requesting users. On the other hand, MF-UAVs ferry and distribute content across the A-UAVs via lateral links. To be noted that that the role of A-UAVs can be served by ground vehicles with similar communication equipment. The rationale behind using such an architecture lies in its ability to provide uninterrupted content provisioning while trying to enhance its caching decisions. The uninterrupted service here is mainly dependent on the A-UAVs, whereas the gradual improvement in the caching decision lies in the collaborative efforts of both A-UAV and swarm of MF-UAVs. The goal is to provide high-availability content access to all the communities without incurring the cost of excessive vertical link usage by the A-UAVs. To achieve this, the paper attempts to answer the following questions. First, what is the benchmark for content caching policies at both A-UAVs and the swarm of MF-UAVs in order to maximize content availabilities to the users. Second, which content should be transferred from A-UAVs to the MF-UAVs to support such caching policies. The proposed policy attempts to address these questions with on-the-fly learning using *Top-k* Multi-armed Bandits.

Currently the UAV-aided Content Delivery Network (CDN) focuses on two primary research directions to tackle the caching problem, namely proactive caching and joint optimization. Note that these terms are broadly used, and they encompass a plethora of approaches which achieves similar goals. Proactive caching techniques like the ones stated in [16]-[18] concentrates on methods that can be used to predict the content demand pattern which can eventually aid in caching decision-making. Such caching techniques are heavily dependent on apriori knowledge of user preferences and content demand patterns. Such assumptions are less likely to work efficiently in ad hoc scenarios like disaster, where the uncertainty in user mobility and surge in dynamic user preferences aggravates the randomness of the content demand pattern.

On the other hand, joint optimization methods generally are centered around multi-objective goals which tend to approach the caching problem along with other objectives [20-26]. Some of the popular problems these methods aim to handle are trajectory planning, information security, eavesdropper detection, energy efficiency, distributed computation, traffic offloading and so on. These approaches are constrained by the prioritization of multiple objectives, which can dilute the focus on optimizing the caching strategy. By embedding the caching problem within a broader set of objectives, these methods inherently limit the algorithm's capacity to achieve an optimal caching policy, as the solution space is divided among competing goals.

From the aspect of performance indicator, content availability serves as a direct indicator of service quality, while user experience is influenced by latency in content access. The central challenge of this paper is to develop an efficient cache placement strategy and content dissemination system capable of adapting to unknown content demand patterns, with the goal of maximizing content availability while controlling access delay. The cache placement strategy must dynamically evolve as user requests are observed, which is crucial for refining caching policies. The effectiveness of the caching decision significantly impacts the placement strategy, yet it must be continuously optimized based on real-time request data.

In this paper, we explore a learning-based approach for content caching within a UAV-assisted dissemination system. Faced with uncertain and heterogeneous content popularity distributions, we formulate the cache placement problem with the objective of maximizing overall content availability. To characterize the uncertainty in content demand, we employ distance-based sequence reordering algorithms. Subsequently, we design a *Top-k* Multi-Armed Bandit learning algorithm for on-the-fly caching. The key contributions of this work are outlined as follows.

- A content dissemination system using swarm of Micro-UAVs is designed for on-demand content dissemination in a communication challenged environment.
- For on-the-fly learning of optimal caching policies, a specific version of bandit problems, namely *Top-k* Multi-armed Bandit is deployed. The proposed system employs Bandit Learning-based caching with a multi-dimensional reward structure in order to learn any inherent patterns in the user content requests as experienced by the micro-UAVs. The learning model within a micro-UAV aims to maximize the cumulative reward [3], which helps in caching decision-making thus improving content dissemination performance.
- A multi-dimensional reward structure for the *Top-k* MAB model is developed based on shared information via micro-UAVs. These rewards take local and global context of content popularities into consideration while learning optimal caching policies.
- A *selective caching* algorithm is designed for joint geographical deployment of Micro-UAVs to manage the trade-off between effective caching capacity and their accessibility.
- The interactions between learnt caching policies and QoS expectation, namely, Tolerable Access Delay, are studied and characterized.
- Simulation experiments and analytical models are developed for functional verification and performance evaluation of the proposed caching and content dissemination framework.

The next section of the paper will discuss the related work on UAV-based caching methods in the existing literature. Following this a concise overview of the system model and the problem definition is discussed. This is followed by a detailed discussion on the baseline cache pre-loading methods. The subsequent section will delve into the proposed learning-based caching policies used for this work. Finally, simulation results are presented to validate the proposed ideas, which helps to draw conclusions from the work done in this paper.

## II. Related Work

In recent years, extensive research has been conducted on UAV-based caching, which can be broadly categorized into platform enhancements and algorithmic optimization. On the platform side, studies such as [4] have demonstrated that the effective caching capacity of UAVs can be significantly increased by leveraging solid-state drives (SSDs) due to their higher storage density and lower power consumption. Similarly, [6] illustrates how extending the communication range between UAVs and ground nodes enhances caching capacity. Additionally, [7] shows that UAVs operating at higher altitudes can cover larger areas, thereby further boosting effective caching capacity. The same study also introduces a multi-UAV caching strategy that optimizes both capacity and coverage by deploying UAVs at different altitudes for specific applications. Energy-efficient algorithms, as explored in [8], [9], employ multi-armed bandit models to optimize user hotspot selection, thereby maximizing data transmission rates while minimizing UAV flight and hovering energy expenditure. However, these approaches often rely on high-performance communication equipment, which increases the payload and accelerates power depletion in UAVs. Furthermore, the high deployment costs associated with larger, high-altitude UAVs with greater payload capacities restrict their population in content dissemination systems. This work addresses these limitations by proposing a system that utilizes a swarm of low-cost micro-UAVs as content carriers across anchor UAVs, which do not require high-performance long-range communication equipment, thereby keeping operational energy budgets low. Moreover, the system's resilience to performance degradation from the loss of a few micro-UAVs enhances its suitability for the proposed content dissemination framework.

From an algorithmic perspective, research such as [14] has explored optimization techniques for flight trajectory, communication scheduling, service coverage extension through optimized UAV hovering time, and multi-hop relaying among UAVs. Similar goals are pursued in IoT networks via multi-hop device-to-device (D2D) routing to extend coverage for energy-constrained UAVs, as detailed in [15]. However, these works generally do not address content placement and caching challenges, which are central to this study.

Specific research on content placement and caching has been undertaken in works like [16]-[19]. For instance, [16] introduces a named data networking (NDN) architecture for IoT networks, where UAVs collect data from the field and deliver it to interested recipients, thus avoiding retransmission. In [17], UAVs proactively transmit content to an algorithmically selected subset of ground nodes, which cooperatively cache the required content. [18] proposes a probabilistic cache placement technique aimed at maximizing cache hit probabilities in networks with wireless nodes distributed according to a homogeneous Poisson Point Process. Meanwhile, [19] focuses on addressing security and denial-of-service attacks in UAV communications. However, these methods do not adequately consider the implications of UAVs' individual and collective storage capacities, making them less suitable for the problem addressed in this paper.

To advance research from an algorithmic standpoint, traffic offloading and learning-based caching strategies have been investigated. For example, [20] demonstrates that considering content popularity and size can significantly enhance the effective caching capacity of UAVs. In [21], a UAV-enabled small-cell network is proposed, where data traffic is offloaded from small-cell base stations (SBSs) to UAVs, with the most popular content proactively cached and delivered directly to users as needed. Similarly, [22] explores the reduction of traffic load on ground base stations through UAV caching. In [23], joint caching and UAV trajectory optimization is achieved using particle swarm optimization, treating each caching strategy as a particle. Furthermore, [24] develops a technique to minimize content delivery delay through joint optimization of UAV trajectory and radio resource allocation, employing deep Q-learning for optimization in large networks with exploding state-action spaces. While these methods address the caching decision problem through traffic offloading and learning-based approaches, they overlook the heterogeneity in content popularity, limiting their effectiveness in scenarios where content demand varies by user geolocation, a key focus of this work.

Additionally, studies such as [25] have explored UAV trajectory control mechanisms that decide whether to continue serving users along a trajectory or return to a charging station based on real-time observations. In a similar joint optimization effort, [26] applies reinforcement learning for UAV caching decisions, where content requests, storage availability, and buffer capacity define states in a Markov Decision Process. These studies, however, do not thoroughly examine the effects of disaster geography, demand heterogeneity, and the impact of UAV trajectories on caching policy, which are the focus of this research.

While certain UAV-based caching mechanisms [21], [22] prove useful in scenarios with partial infrastructure destruction, they are less effective when all communication infrastructures are destroyed, which necessities a fully operational alternative. Moreover, many existing mechanisms [20]-[22] assume temporally static global content popularity, which fails to capture the real-world heterogeneity and time-variability of content demand in disaster scenarios. The optimization techniques employed in [23]-[26], which rely on long-term estimation methods, lack the promptness and adaptability required to respond to changing network and demand conditions. Furthermore, previous work on UAV caching often overlooks the need for effective cache space maximization and reduction of costly vertical link downloads.

To address these gaps, this paper introduces a *Top-k* Multi-Armed Bandit learning model designed for UAV caching decisions, taking into account geo-temporal variations in content popularity and demand heterogeneity. The proposed approach also employs a *selective caching* approach, which enhances system performance by algorithmically leveraging shared information among anchor UAVs and micro-ferrying UAVs.

## III. System Model

In addressing the challenges of dynamic content demand and efficient delivery within disaster-stricken environments, our proposed system leverages a hierarchical UAV architecture designed to optimize content dissemination. The following

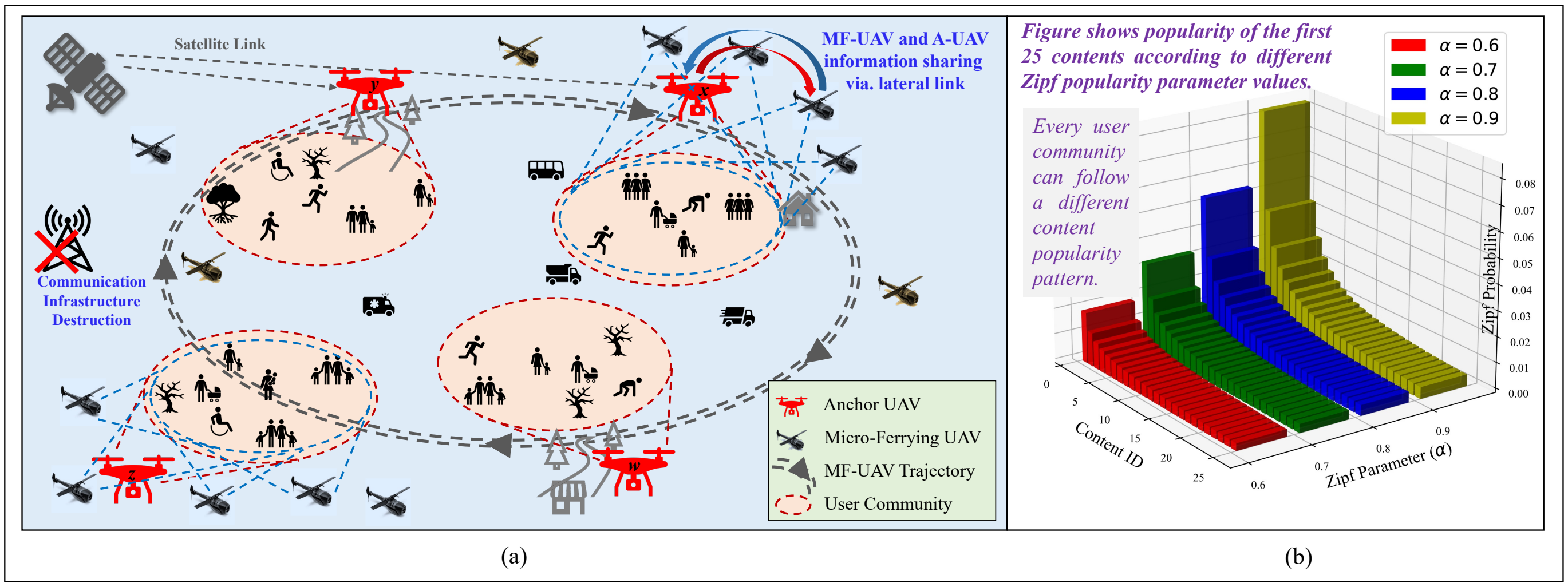

Fig. 1. a) Coordinated UAV system for content caching and distribution in environments without communication infrastructure; b) Zipf Popularity Distribution

section details the underlying system model, including the UAV hierarchy and the mechanisms by which content is managed and provisioned across different user communities.

### A. *UAV Hierarchy*

As shown in Fig. 1, a two-tiered UAV-assisted content dissemination system is deployed. Each community is served by a dedicated A-UAV that uses a lateral wireless connection (i.e., WiFi etc.) to communicate with users in that community. The system introduces a set of low-power-budget Micro-UAVs for the role of ferrying (MF-UAVs). These are unlike A-UAVs which operate with a much larger power budget. MF-UAVs are mobile and possesses only lateral communication links such as Wi-Fi. Unlike the A-UAVs, the MF-UAVs do not possess expensive vertical communication interfaces such as satellite links etc. Effectively, the MF-UAVs act as content transfer agents across different user communities by selectively transferring content across the A-UAVs through their lateral links.

### B. *Content Demand and Provisioning Model*

The content popularity distribution, quality of service and content provisioning are outlined below.

Content Popularity: Research has shown that user content request patterns often follow a Zipf distribution [27], [28]. Here the popularity of a content is proportional to the inverse of its rank, and is a geometric multiple of the next popular content. Popularity of content '$i$' is given as:

$$p_\alpha(i) = \left(\frac{1}{i}\right)^\alpha \Big/ \sum_{k \in N} \left(\frac{1}{k}\right)^\alpha \tag{1}$$

In Equation 1, $\alpha$ is the Zipf parameter that determines the distribution's skewness. The total number of contents in the pool is represented by the parameter $N$. The inter-request time from a user follows the popular exponential distribution [27].

Tolerable Access Delay: For each requested content, the user specifies a Tolerable Access Delay ($TAD$) [4], which serves as a quality-of-service parameter and represents the amount of time the requesting user can wait before the content is downloaded.

Content Provisioning: Upon receiving a request from one of its community users, the relevant A-UAV first searches its local storage for the content. If the content is not found, the A-UAV waits for a potential future delivery by a traveling MF-UAV. If no MF-UAV arrives with the requested content within the specified $TAD$, the A-UAV then proceeds to download it through its vertical link. Since vertical links such as satellite links are expensive, smart caching strategies that can make the content accessible from the UAVs can be effective in reducing content provisioning costs.

Note that Fig. 1 also illustrates the trajectory followed by MF-UAVs, that is the round-robin path, which allows them to have fair access to all communities in their route. This ability allows the content dissemination system to leverage and boost the collective intelligence of the UAV system via collaborative learning. It is worth mentioning that although the system is developed with a trajectory approach that is fair, the developed system has the ability to handle nonuniform trajectories as well. This will be easy to infer as the proposed method and the associated results are discussed. The figure also highlights the coverage of A-UAVs and MF-UAVs. As the choice of content dissemination models dictates that each community is allocated a A-UAV, coverage of a A-UAV is shown through the user community. It has also been shown that the coverage of MF-UAVs only exists when they are in the vicinity of a user community.

Having established the architecture and operational dynamics of the UAV-assisted content dissemination system, it is essential to address the core challenge of optimizing content caching across the UAV network. The following section formulates the caching problem, that defines the objectives and constraints necessary to maximize content availability while adhering to the system's resource limitations and user requirements.

## IV. Content Caching Problem Formulation

For a given number of deployed Anchor and Micro-ferrying

UAVs, the caching problem can be defined as follows.

$$\underset{\forall n \in \mathcal{N}}{maximize} \left[ \frac{1}{\mathcal{N}} \left[ \sum_{n=1}^{\mathcal{N}} P_{avail}(n) \right] \right] \quad (2)$$

$$such\ that \sum_{i=1}^{N_A} C_A + \sum_{j=1}^{N_{MF}} C_{MF} < C \quad (2a)$$

$$and\ \mathcal{T}_{s}^{serve} - \mathcal{T}_{s}^{req} \leq TAD(s), \quad s \in \mathcal{S}_n$$

$$\forall \mathcal{S}_n = \{1,2,3,\cdots\} \quad (2b)$$

Here, $P_{avail}(n) = \frac{\mathcal{S}_n}{\mathcal{R}_n}$, $\mathcal{S}_n$ is the number of contents served at community '$n$' by the UAV/micro-UAV system (both A-UAVs and MF-UAVs), $\mathcal{R}_n$ is the total number of requests made by users at community '$n$', $\mathcal{N}$ is the number of communities, $C$ is the total contents in pool, $C_A$ is the cache of each A-UAV, $C_{MF}$ is the cache of MF-UAVs, $N_A$ is the number of A-UAVs, $N_{MF}$ is the number of MF-UAVs, $\mathcal{T}_{s}^{req}$ is the time at which a content '$s$' is requested by a user, $\mathcal{T}_{s}^{serve}$ is the time when content '$s$' is served to the user by the UAV system, and $TAD(s)$ is the tolerable access delay associated with content '$s$'.

The caching problem focuses on maximizing the overall content availability, as shown in Equation. 2. This objective is constrained by maintaining the cumulative caching capacities of the UAVs below the total number of contents in the content pool, which is captured in Equation. 2a. An additional constraint is imposed by the tolerable access delay associated with a content served to the user by the UAV-aided system (refer to Equation. 2b). In an attempt to solve the aforementioned caching problem, this paper proposes a *Top-k* Multi-Armed Bandit approach that aims to maximize content availability through improvement in the content caching policy.

With the caching problem clearly defined and its constraints outlined, the next step is to explore the baseline caching strategies that can be implemented within the UAV/micro-UAV hierarchy, given an apriori knowledge on content popularity.

## V. Caching based on Content Pre-loading at Anchor UAVs

This section discusses caching policies based on content pre-loading at A-UAVs that assumes pre-assigned, static, and globally known content popularities. After understanding the limitations of these caching policies, the paper proposes a runtime, dynamic, and adaptive *Top-k* Multi-armed Bandit based caching mechanism, which is explained in a Section VI.

### *A. Pre-loading Policies at Anchor UAVs (A-UAVs)*

The *Fully Duplicated* (*FD*) mechanism [27] is a naive approach that allows A-UAVs to download content from vertical links upon request by local users. FD has major limitations including content duplication, high vertical link download costs, and underutilization of UAV cache space. This means that with a cache size of $C_A$ contents per UAV, the total caching capacity of the system is limited to $C_A$. *Smart Exclusive Caching* (*SEC*) [27], [28] overcomes those limitations of FD by storing a set number of unique contents in all A-UAVs and sharing them among communities via traveling MF-UAVs. Assuming globally known *homogeneous content popularity* across all user communities, the *SEC* mechanism divides the cache into two segments of size $C_{S1}$ and $C_{S2}$. Segment-1 contains the top $C_{S1} = \lambda C_A$, popular contents cached in all A-UAVs, while Segment-2 contains unique contents $C_{S2} = (1-\lambda)C_A$, where $\lambda$ is a *Storage Segmentation Factor*. This results into $C_{S2}^{total} = N_A(1-\lambda)C_A$, which is the total number of Segment-2 contents stored across all $N_A$ number of A-UAVs, and these can be shared across all user communities via the mobile MF-UAVs. This factor needs to be adjusted and fine-tuned based on various network, content, and demand conditions. Total number of contents in the system as per *SEC* is given as:

$$C_{sys} = \lambda C_A + N_A(1-\lambda)C_A \quad (3)$$

*Popularity-Based Caching* (*PBC*) [29] is employed when different communities have different content preferences. Considering the *heterogeneous popularity* sequence of a community, the *PBC* approach, like *SEC*, divides the cache space of the local A-UAV into two segments of size $C_{S1}$ and $C_{S2}$. Segment-1 caches the most popular contents, which can be exclusive to a A-UAV ($C_E$) or non-exclusive i.e., may be cached across multiple A-UAVs ($C_{NE}$), such that, $C_{S1} = C_E + C_{NE}$. To be noted that according to the exclusivity of contents in $C_{S1}$, the total number of exclusive contents across all A-UAVs is termed as $C_E^{total}$. This leads to more diverse contents being stored in Segment-1 across all A-UAVs. Segment-2 is the same as that in *SEC*. Therefore, by modifying Equation 3, the total number of contents in the system can be expressed as:

$$C_{sys} = C_{NE} + C_E^{total} + N_A(1-\lambda)C_A$$

$$\Rightarrow C_{sys} \geq \lambda . C_A + N_A(1-\lambda)C_A \quad (4)$$

*Value-Based Caching* (*VBC*) [29] further enhances the caching policy by storing top-valued contents in Segment-1 of the A-UAVs, where *value* of contents comprises of their popularity and tolerable access delay. Value of a content '$i$' is calculated as:

$$V(i) = \kappa \upsilon^* \left( \frac{p_\alpha(i)}{TAD(i)} \right)$$

$$= V(i) = \kappa \left( \frac{TAD_{min}}{p_\alpha(1)} \right) \left( \frac{p_\alpha(i)}{TAD(i)} \right) \quad (5)$$

In this equation, $p_\alpha(i)$ represents the content's popularity as per the Zipf distribution, $TAD(i)$ is the content's tolerable access delay, $\kappa$ is a scalar weight that increases as popularity decreases, and $\upsilon^*$ is a normalization constant. The normalization constant is calculated for a given Zipf (popularity) parameter $\alpha$ using the minimum possible $TAD$ ($TAD_{min}$ ) and the maximum possible popularity, which is $p_\alpha(1)$, i.e., $\upsilon^* = TAD_{min}/p_\alpha(1)$. The value of $V(i)$ is bounded between [0,1], and it increases as $p_\alpha(i)$ increases and $TAD(i)$ decreases. The content's value presents a holistic quantifiable

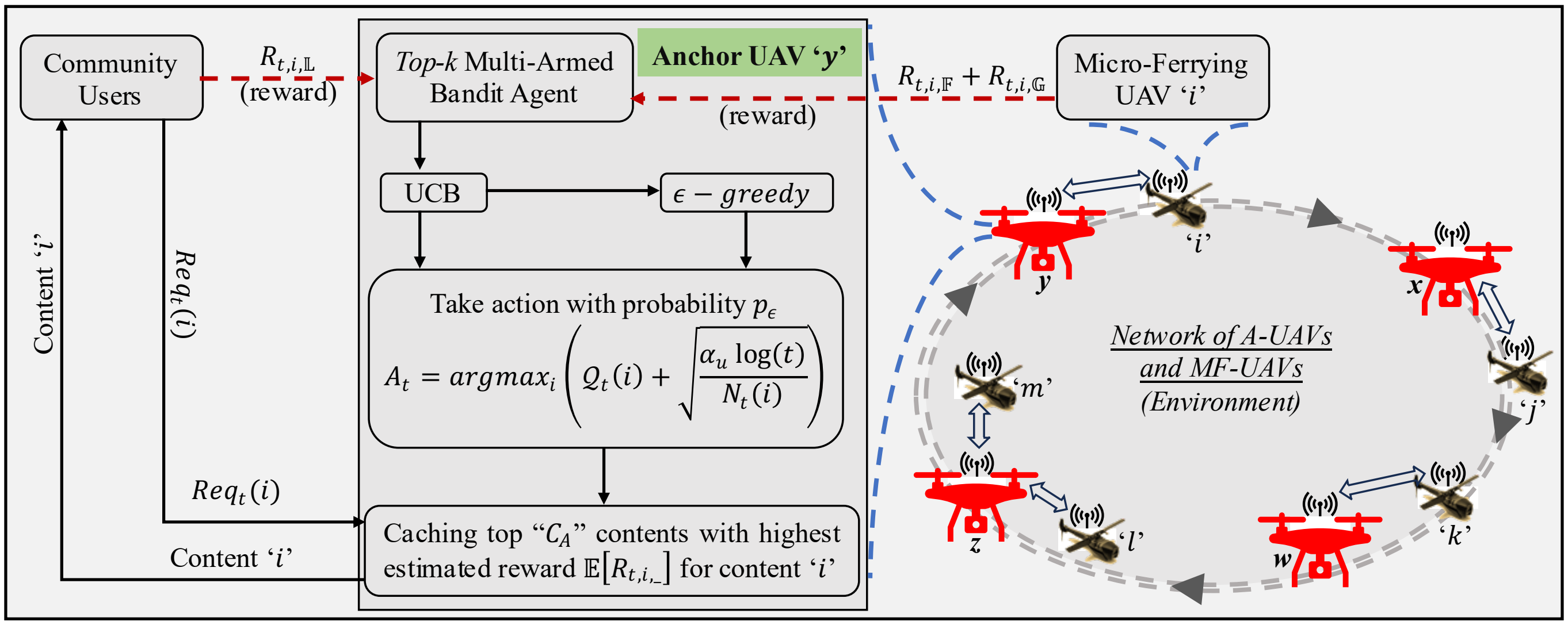

Fig. 2. *Top-k* Multi-Armed Bandit Learning for Caching Policy at A-UAVs

measure for caching decision.

The caching policy for micro-ferrying UAVs remains the same for all the above-discussed caching policies for A-UAVs, which will be discussed in the forthcoming Section VI. An MF-UAV ferries content across the A-UAVs it visits along its trajectory. The caching policy of A-UAVs determines the utility of MF-UAVs where every A-UAV should maintain sufficient contents in its cache space to maximize the MF-UAV cache space utilization.

### B. Limitations of Cache Pre-loading at A-UAVs

The caching policies discussed in this section rely on pre-loading content into A-UAVs, which has certain limitations. These approaches assume *a priori* knowledge of the popularity distribution of all the content in the system, which can hinder practical feasibility during deployment. Local popularity estimation of requested content within individual A-UAVs can partially alleviate this issue, but it cannot adjust the crucial storage segmentation factor ($\lambda$) (see Section V) for maximizing availability across the entire system of A-UAVs and their communities. Collaborative global popularity estimation can be introduced, but it fails to capture locally meaningful demand heterogeneity across different communities.

This exploration leads us toward developing a more adaptive and responsive caching policy, capable of dynamically learning and adjusting to the varying user preferences in real-time. In the next section, we delve into a decentralized caching approach utilizing *Top-k* Multi-Armed Bandit learning, designed to address the identified shortcomings and enhance the system's adaptability to diverse user demand patterns.

## VI. Decentralized Caching with Multi-Armed Bandit

This section presents a plausible solution for the aforementioned shortcomings by using *Top-k Multi-Armed Bandit* learning for caching decisions at the A-UAVs. This facilitates faster learning and is adaptive to heterogeneous user demand patterns through information sharing via micro-UAVs. Based on the forthcoming mechanism, the caching policy for micro-ferrying UAVs is also modified to leverage their ubiquity, which is discussed later.

### A. Top-k Multi-Armed Bandit Learning

Multi-Armed Bandit is a classic problem in reinforcement learning [3] and decision-making. At each round $t$, an agent chooses an arm $A_t$ out of $N$ arms, denoted by $A_1, A_2, \ldots, A_N$, and observes a reward $R_t$. Each arm $i$ has an unknown reward distribution with mean $\mu_i$ and variance $\sigma_i^2$. The agent's goal is to maximize the total expected reward $R_T$ over $T$ rounds, where $T$ is the total number of rounds (time horizon):

$$R_T = max \sum_{t=1}^{T} E[R_t] \tag{6}$$

This paper uses a variant of MAB called *Top-k Multi-Armed Bandit* [30]. Here, the agent has to choose $k$ arms simultaneously out of a larger set of $N$ arms, and it receives a reward for each arm in the chosen set. This is in contrast to choosing only one arm in classical MAB approaches. The goal of the agent is to maximize the total cumulative reward $R_T$ obtained over a finite time horizon $T$:

$$R_T = max \sum_{t=1}^{T} \sum_{i=1}^{k} E[R_{t,i}] \tag{7}$$

### B. Caching at A-UAV using Top-k Multi-Armed Bandit

In the scenario of UAV-caching, there is a *Top-k* MAB agent in each A-UAV. Here, choosing each content for caching corresponds to choosing an arm. The '$k$' of *Top-k* MAB agent corresponds to the caching capacity of A-UAV, i.e., $k = C_A$. The agent's aim is to select '$C_A$' contents out of the total pool of '$N$' contents to be cached in an A-UAV such that the content availability to the users can be maximized. Here, the UAV-aided content dissemination system is the learning environment where the A-UAVs interact through their actions of choosing specific sets of contents to be cached. The feedback from the environment for the taken actions are in the form of rewards/penalties. Micro-ferrying UAVs play a crucial role in transferring information across the UAV-aided system, which helps in the computation of appropriate rewards/penalties, as shown in Fig. 2. Actions are rewarded when cached contents are requested by the users and are served to the users within the given tolerable access delay or penalized otherwise. The top $C_A$ contents that accumulate most reward from the corresponding community and other communities are chosen to be cached at a A-UAV. It should be noted that the *Top-k* MAB agents in the A-UAVs are provided with no *a priori* information about the content popularity at the corresponding user communities.

A good choice for learning decision epoch in each *Top-k* MAB agent is according to the MF-UAVs accessibility at the corresponding community (i.e., an MF-UAV's visiting frequency). This is because the MF-UAVs carry the content availability information from the communities in its trajectory. Such information is leveraged for learning at the A-UAVs' *Top-k* MAB agents using appropriately designed multi-dimensional rewards. The agent learns to cache contents via the multi-dimensional reward structure which has three parts, namely, *local*, *ferrying,* and *global* rewards. These are the components of the reward mentioned in Equation 7 and are explained next. Let $\mathbb{L}_t$, $\mathbb{F}_t$ and $\mathbb{G}_t$ denote the sets of locally requested contents, contents requested at other communities, and contents requested across all communities at $t^{th}$ epoch, respectively. These contents can be served to the users directly by a A-UAV or indirectly via the visiting MF-UAVs. If a cached content is served to a user within the given *TAD* and an increase in content availability is observed, the content is rewarded. The type of reward is determined by the set to which the cached content belongs. The expressions for three types of rewards, at $t^{th}$ epoch, are given as follows:

$$R_{t,i,\mathbb{L}} = \mathbb{I}_1\left(i \in \mathbb{L}_t, \delta_{t,\mathbb{L}} \geq 0\right) + \mathbb{I}_{-1}\left(i \notin \mathbb{L}_t, \delta_{t,\mathbb{L}} < 0\right) \quad (8)$$

$$R_{t,i,\mathbb{F}} = \frac{1}{N_A - 1} \sum_{j=1, j\neq \mathbb{X}}^{N_A} \mathbb{I}_1\left(i \in \mathbb{F}_t, \delta_{t,\mathbb{F}} \geq 0\right) + \frac{1}{N_A - 1} \sum_{j=1, j\neq \mathbb{X}}^{N_A} \mathbb{I}_{-1}\left(i \notin \mathbb{F}_t, \delta_{t,\mathbb{F}} < 0\right) \quad (9)$$

$$R_{t,i,\mathbb{G}} = \frac{1}{N_A} \sum_{j=1}^{N_A} \mathbb{I}_1\left(i \in \mathbb{G}_t, \delta_{t,\mathbb{G}} \geq 0\right) + \frac{1}{N_A} \sum_{j=1}^{N_A} \mathbb{I}_{-1}\left(i \notin \mathbb{G}_t, \delta_{t,\mathbb{G}} < 0\right) \quad (10)$$

$$where, \mathbb{I}_1(A) = \begin{cases} 1, & if\ A\ is\ true \\ 0, & otherwise \end{cases}$$

The above equations compute the reward at $t^{th}$ epoch according to increase in availability due to content '$i$' cached at A-UAV '$\mathbb{X}$'. Here, $R_{t,i,\mathbb{L}}$, $R_{t,i,\mathbb{F}}$, and $R_{t,i,\mathbb{G}}$ are local, ferrying, and global rewards respectively. The terms $\delta_{t,\mathbb{L}}$, $\delta_{t,\mathbb{F}}$ and $\delta_{t,\mathbb{G}}$ correspond to the increase in local availability, ferried content availability and global availability respectively. Each type of reward is contingent upon the condition in the indicator function $\mathbb{I}_{1/-1}(i)$. The first terms in Equations. 8, 9 and 10 represent the reward accumulated by caching content '$i$' at A-UAV '$\mathbb{X}$', whereas the second term is the penalty associated with adverse condition. To be noted that $R_{t,i,\mathbb{F}}$, and $R_{t,i,\mathbb{G}}$ are higher if the content '$i$' is requested and served at more communities.

Learning is achieved using a tabular method where a Q-table is maintained for all contents in the A-UAVs. The value corresponding to each content is called a Q-value or action-value [3]. The agent updates the Q-value for a content at every learning epoch according to the multi-dimensional rewards in Equations. 8-10 from the interaction with the environment (UAV-aided content dissemination system) and learns the best actions (contents cached). The recursive expression which explains Q-value update for a content '$i$' at A-UAV '$\mathbb{X}$' is given as follows:

$$\mathcal{Q}_{t+1}(i) = (1-\eta)\mathcal{Q}_t(i) + \eta\left(R_{t,i,\mathbb{L}} + \mathbb{I}_1(\delta)\left(R_{t,i,\mathbb{F}} + R_{t,i,\mathbb{G}}\right)\right) \quad (11)$$

Here, $\mathcal{Q}_t(i)$ represents the Q-value of a content '$i$' at $t^{th}$ epoch; $R_{t,i,_}$ is the respective reward received by caching content '$i$'; $\delta$ represents the condition for the indicator function $\mathbb{I}_1(\delta)$ which is 1 if micro-ferrying UAVs are present in the communication range of A-UAV '$\mathbb{X}$' or 0 otherwise; $\eta$ is a hyper-parameter which controls the learning rate. The Q-values for all contents are initialized with zero to ensure no *a priori* information for a *Top-k* MAB agent. Also, it ensures equal importance to all contents for caching decisions. As learning progresses, Q-values improve and best contents with highest Q-values are cached with the aim of maximizing accumulated reward which improves the caching policy and thus increases content availability.

Note that there can be very large number, i.e., $\binom{N}{k}$, of combinations of contents to be sampled by the *Top-k* MAB agent for caching. Consequently, the reward estimation for each individual content combination occurs infrequently, only after large intervals. This can lead to a weak estimates of reward distribution, as the global content population size $N$ increases. This issue is handled by empirically selecting $\epsilon$ and its decay rate in the $\epsilon$-greedy action selection policy [30]. To reduce the dependence of a caching policy on the choice of $\epsilon$, an Upper Confidence Bound (UCB) strategy is used [30]. The *Top-k* MAB agent maintains an upper confidence bound on the expected reward of each content, and selects the set of $C_A$ contents with the highest UCB at each epoch.

$$\mathcal{U}_t(i) = \mathcal{Q}_t(i) + \sqrt{\frac{\alpha_u \log(t)}{N_t(i)}} \quad (12)$$

Here, $\mathcal{U}_t(i)$ is the UCB of content '$i$' at epoch '$t$'; $\mathcal{Q}_t(i)$ is the updated Q-value at epoch '$t$'; $\alpha_u$ is a hyperparameter that controls the degree of exploration; $N_t(i)$ is the number of time content '$i$' has been requested till epoch '$t$'. The first term represents the reward estimate, and the second term depicts the uncertainty in reward estimate. UCB selects the content that has high potential for high reward but hasn't been requested frequently. This promotes exploration without externally inducing an exploration parameter such as $\epsilon$. For this paper, $\mathcal{U}_t(i)$ is used in place of $\mathcal{Q}_t(i)$ to cache content '$i$', as shown in Step 7-14 in Algorithm 1.

The following pseudo code explains the caching policy at a micro-ferrying UAV with a *Top-k* MAB agent.

**Algorithm 1** Caching policy at a A-UAV with *Top-k* MAB Learning

1. **Initialization:**
   a. N: Total contents in the system
   b. $C_A$: Caching capacity of an A-UAV
   c. $\mathcal{U}$: Size $|C_A|$ initialized with 0's (Q-table with UCB)
   d. $\eta$: Learning rate for Q-table update
   e. $\alpha_u$: Degree of exploration (in UCB)
2. **Load** A-UAV's cache with $C_A$ randomly chosen contents.
3. **while** True:
4.   **Check** for learning epoch at A-UAV i.e., at $t^{th}$ epoch
5.   **if** True **then do**
6.     **for** $i = 0$ $to$ $length$(A-UAV cache size $C_A$) **do**
7.       **Get** reward $R_{t,i,_}$ \\ according to Equations. 8-10
8.       **Update** $\mathcal{U}(i)$ \\ from Equations. 11 and 12
9.     **end for**
10.     $value = \boldsymbol{copy}(\mathcal{U})$ \\ make a copy of UCB values \\ Reload contents (Select arms)
11.     **for** $i = 0$ $to$ $length$(A-UAV cache size $C_A$) **do**
12.       $c_{max} = \boldsymbol{argmax}(value)$
13.       **Load** $c_{max}$ to A-UAV
14.       **Set** $value[c_{max}] = -\infty$
15.     **end for**
16.   **end if**
17. **end while**

### C. *Proof of convergence*

Within a finite time horizon, the *Top-k* MAB agent at a A-UAV converges to a caching policy which approaches the benchmark caching policy asymptotically. The proof of convergence lies in the intrinsic regret minimizing characteristics of MAB [3], which is shown below.

$$C_A = \{i | i \in N, 1 \le i \le k\} = \underset{k}{\operatorname{argmin}}\big(Regret(T)\big)$$
$$= \underset{k}{\operatorname{argmin}}\left(\sum_{t=1}^{T}\left(\max_k \sum_{i=1}^{k} R_{t,i^*} - \sum_{i=1}^{k} R_{t,i}\right)\right) \quad (13)$$

where, $T$ is the total number of epochs (time horizon); $k$ is the number of contents cached at each epoch; $i^*$ represents the optimal caching action; $i$ is the caching action selected by the *Top-k* MAB agent at $t^{th}$ epoch. Equation. 13 shows the difference between the reward obtained by the algorithm and the reward obtained by caching with benchmark policy. Post-convergence, the instantaneous regret should be minimum, which is experimentally proven in this paper. Ideally for a perfectly designed reward structure the regret should asymptotically vanishes, i.e., $\lim_{T\to\infty}\frac{Regret(T)}{T} = 0$ [31].

The convergence of estimated rewards (Q-values) to the true values (expected reward) in a MAB setup, including *Top-k* MAB scenarios, can be analyzed using the Law of Large Numbers (*LLN*) [32] and concepts of stochastic approximation. For simplicity, this work initially considers the proof for a single arm and then extend the idea to all '$k$' arms in the *Top-k* selection. According to weak law of large numbers [32], the estimated value of a content '$i$' will be at a minute offset '$\epsilon_i$' from its true value, which is shown in the following expression:

$$\left|\lim_{T\to\infty} \mathcal{Q}_{t+1}(i)\right| - \mu_i^* < \epsilon_i$$
$$= \left|\lim_{T\to\infty} \frac{1}{n}\sum_{t=1}^{n}\left[R_{t,i,\mathbb{L}} + \mathbb{I}_1(\delta)\big(R_{t,i,\mathbb{F}} + R_{t,i,\mathbb{G}}\big)\right]\right| - \mu_i^* < \epsilon_i \quad (14)$$

Here, a single content/arm '$i$' has a true value of $\mu_i^*$ , and $\mathcal{Q}_{t+1}(i)$ represent the estimated reward (Q-value) of content '$i$' after it has been selected '$n$' times. The reward is taken from the second term (weighted reward) of Equation. 11. For convergence, the weight '$\eta$' is chosen empirically in such a way that it satisfies the *Robbins-Monro stochastic approximation condition* [31] for non-constant '$\eta$', namely $\sum_n \eta_n(i) = \infty$, and $\sum_n \eta_n(i)^2 < \infty$. To be noted that the weight '$\eta$' is manifestation of '$1/n$' in Equation. 14. Now, extending the concept to all top '$k$' contents, Equation. 14 can be modified using Equation. 7:

$$\left|\lim_{T\to\infty} \frac{1}{n}\sum_{t=1}^{n}\left[\sum_{i=1}^{k} \mathcal{Q}_{t+1}(i)\right]\right| - \sum_{i=1}^{k}\mu_i^* < \sum_{i=1}^{k}\epsilon_i$$
$$\Rightarrow \left|\lim_{T\to\infty} \frac{1}{n}\sum_{t=1}^{n}\left[\sum_{i=1}^{k}\left[R_{t,i,\mathbb{L}} + \mathbb{I}_1(\delta)\big(R_{t,i,\mathbb{F}} + R_{t,i,\mathbb{G}}\big)\right]\right]\right|$$
$$- \sum_{i=1}^{k}\mu_i^* < \sum_{i=1}^{k}\epsilon_i \quad (15)$$

The convergence proof for each of the top '$k$' contents individually follow the same logic as for the single content, provided each content is sampled infinitely often. Each content, including the top '$k$' contents, must be selected infinitely often as the number of total selections $T \to \infty$. This requirement is met in practice by exploration strategies (like $\epsilon$-greedy/UCB) that ensure all arms are explored sufficiently over time.

With an assumption on the success of the *Top-k* MAB based caching policy, let's say that the ideal sequence of contents are cached at A-UAVs, which is $C_A = \{i^* | i^* \in N, 1 \le i^* \le k\}$. For this caching decision, $\sum_{i=1}^{k}\epsilon_i = 0$, according to the expression given in Equation. 15. Therefore, the instantaneous regret post-convergence can be derived from Equation. 13 and 15, as follows:

$$\max_k \sum_{i=1}^{k}\left[R_{t,i^*,\mathbb{L}} + \mathbb{I}_1(\delta)\big(R_{t,i^*,\mathbb{F}} + R_{t,i^*,\mathbb{G}}\big)\right]$$
$$- \sum_{i=1}^{k}\left[R_{t,i,\mathbb{L}} + \mathbb{I}_1(\delta)\big(R_{t,i,\mathbb{F}} + R_{t,i,\mathbb{G}}\big)\right] \approx 0 \quad (16)$$

The evidence of convergence, supporting the above expression is shown in Fig. 6, where near-optimal contents cached at A-UAVs leads to $\sum_{i=1}^{k} \epsilon_i \approx 0$. According to the learnt caching policy, the cached contents can boost content availability at their respective communities as well as at other distant communities via MF-UAVs.

*D. Selective Caching at Micro-Ferrying UAVs (MF-UAVs)*

Ideally, the purpose of the MF-UAVs is to ferry around a subset of $C_E^{total} + N_A(1-\lambda)C_A$ number of contents stored across $N_A$ number of A-UAVs (see Section V). Due to the limitation of per-MF-UAV caching space (i.e., $C_{MF}$), its caching policy should be determined based on its trajectories, learnt caching policy at A-UAVs, content request patterns, and the $TADs$ associated with the contents to be cached.
MF-UAV caching policy is explained in the pseudocode below.

**Algorithm 2** MF-UAV Caching Algorithm with *Top-k* MAB learning-based caching policy at A-UAVs

**Input:** Total A-UAVs in its trajectory, $TAD$, next A-UAV '$x$', present A-UAV '$x-1$'
**Output:** $C_{MF}$ contents for MF-UAV '$y$'
Caching at A-UAVs with *Top-k* MAB policy (*Algorithm 1*)
**while** True:
  **if** MF-UAV leaving for next A-UAV '$x$' **then do**
    // Contents that are not in the future visiting A-UAV
    **Update** ferrying content knowledge
    // Function call from the present A-UAV '$x-1$'
    **Call** content-wise_TAD ( )
    // Present A-UAV sends MF-UAV visiting frequency
    **Call** MF-UAV_visiting_frequency ( )
    // Check what content the last MF-UAV ferried
    **Call** Check_previous_MF-UAV_roster ( )
      **Return** roster contents with respective TADs
    // Compute request interval for last MF-UAV roster
    **Calculate** least popular content's request interval
    **Check** if request time is less than its TAD and MF-UAV visiting duration
    **if** True **then do**
      Cache same roster
    **else**
      Cache next best roster
    **end if**
    **Check** if other MF-UAVs flying with MF-UAV '$y$'
    **for** $l = 0$ to $length$(MF-UAVs flying together) **do**
      **for** $k = 0$ $to$ $length$(A-UAV '$x$' cache $C_A^x$) **do**
        **Check** if $k$ in $C_{MF}$ cache space of MF-UAV '$y$'
        **if** True **then do**
          Replace '$k$' with highest value content from $C_A^{x-1}$ not cached in MF-UAV '$y$' and A-UAV '$x$'
        **end if**
      **end for**
      Cache next best roster
    **end for**
  **end if**
  **Update** next A-UAV '$x$', present A-UAV '$x-1$'
**end while**

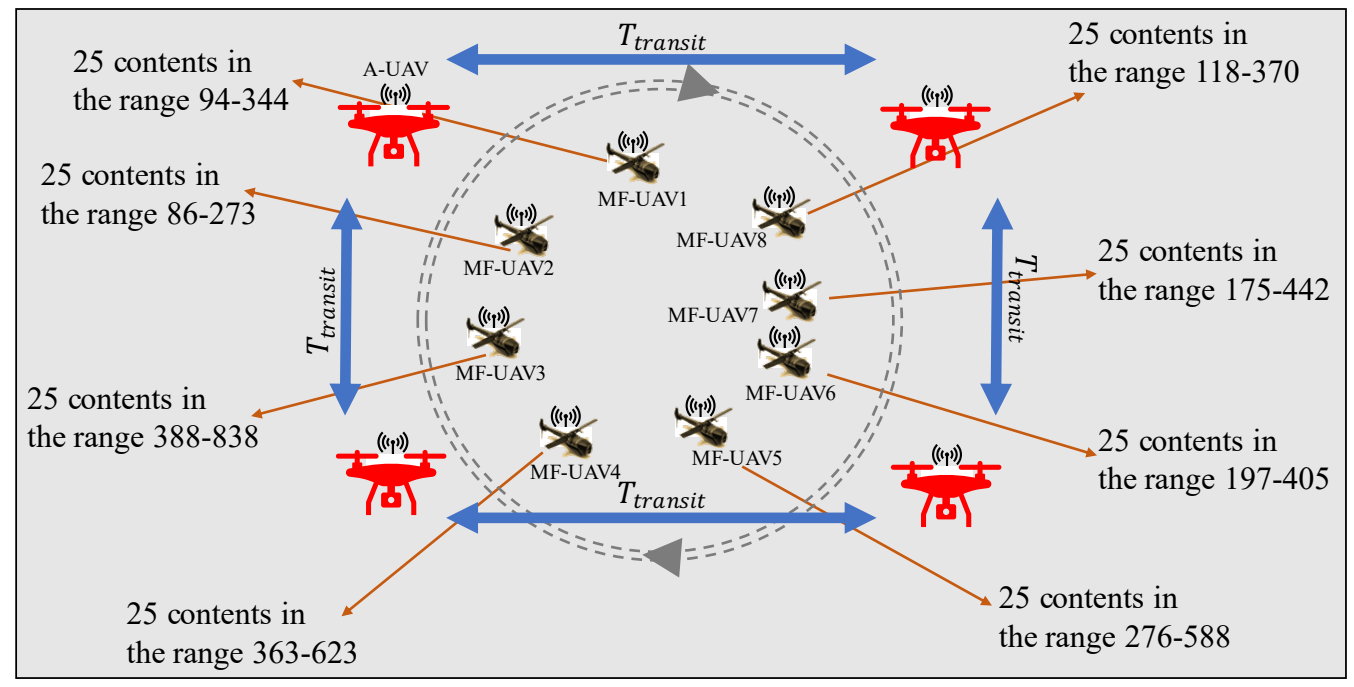

Fig. 3. Algorithmic selection of cached contents at MF-UAVs in conjunction with *Top-k* Multi-Armed Bandit learning at A-UAV

The role of MF-UAVs is to ferry contents from the previously visited A-UAVs to the future visiting A-UAV such that the future visiting A-UAV gets the benefit of contents cached at other A-UAVs. In Algorithm 2, this process is described in detail. Fig. 3 shows the impact of this collaborative algorithm.

Consider a situation in which an MF-UAV '$y$' is ready to leave the A-UAV '$x-1$'. Before caching contents, it needs the following information from A-UAV '$x-1$'; 1) What are the contents eligible for ferrying; 2) What is the MF-UAVs visiting frequency; 3) What roster of ferrying content did the last MF-UAV ferry, where roster is the grouping of contents based on their popularity or value; 4) Are the next roster contents likely to be requested within the given *TAD*; and 5) Are MF-UAVs flying in close proximity with each other. Based on these information MF-UAV '$y$' selectively caches contents while maintaining diversity in the contents cached by other MF-UAVs in its proximity. This means, if MF-UAVs are flying while maintaining proximity with each other or in groups, they ferry contents from consecutive rosters. To be noted that the size of a roster is same as an MF-UAV's cache size. Therefore, if MF-UAVs are flying in groups of $N_{MF}^G$ (group size), then the number of contents cached by the group is $N_{MF}^G C_{MF}$. Such selective caching policy at MF-UAVs ensures content availability maximization by avoiding redundant cache duplication.

*E. Complexity Analysis*

To evaluate the computational feasibility of the *Top-k* Multi-Armed Bandit algorithm, we analyzed the complexity of its key components, including initialization, reward calculation, action selection, and the Q-value updates. These components collectively determine the scalability of the algorithm for large-scale UAV-assisted content dissemination systems.

The algorithm begins with an initialization phase where each A-UAV crates a Q-table to track the rewards associated with all $N$ contents. This initialization step, which is performed only once, has a time complexity of $O(N)$. Given the one-time nature of this step, its computational burden does not affect the scalability of the system in a major way. During each learning epoch, the algorithm computes rewards from $k$ arms selected from the total $N$ contents. This reward calculation, which occurs across $T$ epochs, involves evaluating the multi-dimensional reward structure. As $k \ll N$, the complexity of this step is $O(T.k)$, making it computationally light enough for real-time operations. The most computationally intensive step is the action selection process, where the Upper Confidence Bound (UCB) method ranks all $N$ contents to identify the top $k$ arms to cache. Sorting the contents at each epoch results in a complexity of $O(N \log N)$, which, over $T$ epochs, leads to a total complexity of $O(T.N \log N)$. While this step dominates the overall computational complexity, it is till bounded by logarithmic growth, thus keeping it scalable for large content pools. Finally, the Q-values of the selected arms are updated at each epoch based in the observed rewards. This update process has a complexity of $O(T.k)$, which remains manageable due to the small value of $k$ relative to $N$.

Combining these components, the overall time complexity of the *Top-k* MAB algorithm is $O(N) + O(T.k) + O(T.N \log N) + O(T.k)$. Simplifying this for $k \ll N$, the dominant term is $O(T.N \log N)$. Therefore, the overall time complexity is $O(T.N \log N)$, with the action selection step contributing the most significant computational cost. This complexity demonstrates the algorithm's suitability for large-scale systems, as its logarithmic scaling ensures efficiency even with large number of contents. Furthermore, the selective caching mechanism employed by MF-UAVs, with a linear complexity of $O\left(N_{Trajectory}.(C_{MF} + N_A)\right)$, complements the learning-based approach by ensuring effective content distribution without introducing significant computational overhead. Note that here $C_{MF}$ is the cache size of MF-UAVs, $N_A$ is the number of A-UAVs, and $N_{Trajectory}$ is the number of trajectory changes.

Thus, the overall complexity of the *Top-k* MAB algorithm is dominated by $O(T.N \log N)$, which makes it scalable for large content pools. The linear complexity of the MF-UAV caching mechanism further supports efficient operation in distributed settings. The proposed framework is computationally efficient, scalable and is well-suited for real-time UAV-assisted content dissemination in disaster-stricken environments.

## VII. Experimental Results and Content Dissemination Performance

Simulation experiments are performed to analyze the performance of the proposed *Top-k* MAB learning-based caching mechanism and selective caching at the micro-ferrying UAVs. An event-driven simulator accomplishes content request generation while maintaining an intra-event interval according to exponential distribution and following a Zipf popularity distribution (refer Equation. 1). To capture heterogeneity in content popularity sequence at different communities, contents are swapped with pre-decided probability [29] and the difference between the sequences are determined using Smith-Waterman Distance [29]. The default experimental parameters for the proposed *Top-k* MAB learning based caching and cache pre-loading policies are listed in Table I.

TABLE I.
Default Values for Model Parameters

| # | Variables | Default Value |
|---|---|---|
| 1 | Total number of contents, $C$ | 2000 |
| 2 | Number of A-UAVs, $N_A$ | 4 |
| 3 | Number of MF-UAVs, $N_{MF}$ | 8 |
| 4 | A-UAV's Cache space (content count), $C_A$ | 200 |
| 5 | MF-UAV's Cache space, $C_{MF}$ | 25 |
| 6 | Poisson request rate parameter, $\mu$ (request/sec) | 1 |
| 7 | Hover rate of MF-UAV, $R_{Hover} = T_{Hover}/T_{Trajectory}$ | $1/6$ |
| 8 | Transit rate of MF-UAV, $R_{Transit} = T_{Transit}/T_{Trajectory}$ | $1/12$ |
| 9 | Zipf parameter (Popularity), $\alpha$ | 0.4 |
| 10 | Micro Ferrying UAV Trajectory | Round-robin |

In the simulation, the impact of lateral link range on content dissemination has been implemented. An MF-UAV begins serving content upon entering the WiFi transmission range of a community, even before reaching its boundaries. The duration during which the MF-UAV starts transmitting content, denoted as $\Delta t_{comm}$, is influenced by its transit speed. If $\Delta t_{comm}$ is significantly shorter than the Poisson-distributed content request generation time ($T_{req}$), the adjusted hover time remains approximately the same ($\hat{T}_{Hover} \approx T_{Hover}$). Conversely, if $\Delta t_{comm}$ is comparable to or exceeds $T_{req}$, the adjusted hover time increases to $\hat{T}_{Hover} \approx T_{Hover} + \Delta t_{comm}$, while the transit time decreases to $\hat{T}_{Transit} \approx T_{Transit} - \Delta t_{comm}$.

The performance evaluation of the proposed mechanism is accomplished via the following metrics.

*Content Availability* ($P_{avail}$): Defined as the ratio between cache hits and generated requests for a given tolerable access delay. Cache hits are the content provided to the users from the contents cached in the UAV-aided caching system (without download). Therefore, content availability indirectly indicates the content download cost of a systems as well.

*Cache Distribution Optimality* (*CDO*): This determines the optimality of the learnt caching policy in terms of the caching sequence. Jaro-Winkler Similarity (*JWS*) [29] is used to represent *CDO*, by computing the similarity between the content sequence from the learnt caching policy and content sequence according to cache pre-loading. It is computed by calculating the number of matches, number of transpositions required within the matches and the similarity in prefix of both sequences. It is a normalized similarity measure where 1 represents optimal caching and 0 means non-optimal caching.

*Access Delay* (*AD*): Performance of *Top-k* MAB model and selective caching policy for micro-ferrying UAVs is also evaluated based on the access delay which is the end-to-end delay between the generation of content request and its provisioning from the cached contents in the UAVs. This paper reports the epoch-wise average access delay to show the improvement in caching policy as learning progresses.

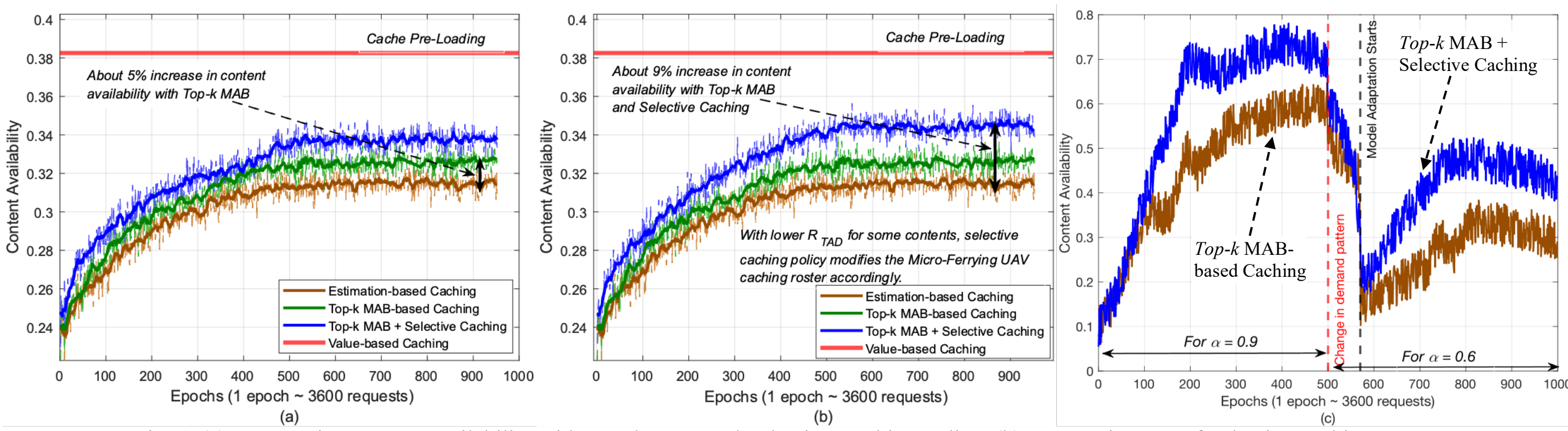

Fig. 4. (a) Increase in Content Availability with *Top-k* MAB and Selective Caching Policy, (b) Responsiveness of Selective caching to user demand i.e., TAD, (c) Reactive nature of the proposed caching policy in case of time-varying user preferences

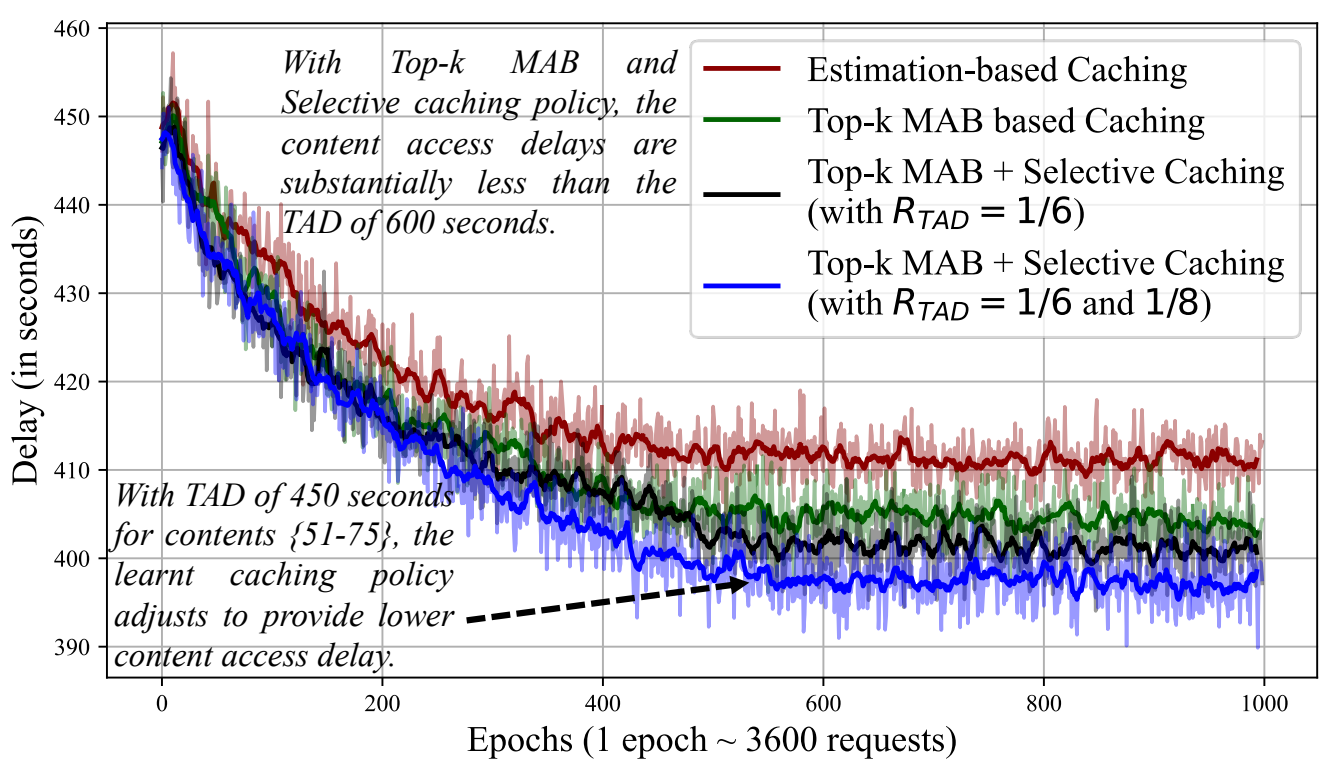

Fig. 5. Delay with *Top-k* MAB and Selective Caching Policy

### A. *Effect of Exploration Strategies on Learnt Caching Policy*

In order to understand the viability of the proposed *Top-k* MAB learning-based caching policy in scenarios with demand heterogeneity, two type of content popularity sequence are used. This is achieved with adjacent communities having different popularity sequences. For UCB exploration strategy, the degree of exploration is set to $\alpha_u = 2$. Also, to show the effectiveness of selective caching at micro-ferrying UAVs (MF-UAVs), *TAD* Ratio $R_{TAD}$ for contents $\{51 - 75\}$ are kept lower than the default $R_{TAD}$ i.e., $1/8$ . The choice of *TAD* for the experiments is such that it is less than the hovering and transiting duration together. This is done to emphasize the reactive nature of the algorithm by constraining the allowed duration for a request, before it is served via download. It should be highlighted that keeping the *TAD* too high allows the MF-UAVs to reduce the caching frequency of those contents. On the contrary, for very low *TAD*, the model overestimates the value of those contents leading to them being cached at A-UAVs allowing ready availability. To be noted that *TAD*s are represented as a ratio with respect to trajectory time ($T_{Trajectory}$) to ensure generalizability of the proposed algorithms. Fig. 4 shows the convergence behavior of the learnt caching policy with *Top-k* MAB model at the A-UAVs, and selective caching at the MF-UAVs.

The convergence behavior is shown in terms of content availability from the learnt caching policy. The observations from Fig. 4 are as follows. First, the figure shows that by employing *Top-k* MAB agent at every A-UAV and selective caching at MF-UAVs, a caching policy can be learnt which can provide content dissemination performance closer to the benchmark performance [29]. The benchmark performance, using Value-based Caching as shown in Fig. 4, is calculated with the aid of *apriori* information on content popularity and takes into consideration the heterogeneity in user demand. The proposed *Top-k* MAB algorithm is able to leverage the multi-dimensional reward structure, as explained in Equations. 8-10, to learn the caching policy on-the-fly (see Section VI). Second, the selective caching policy at micro-ferrying UAVs leverages the shared information between themselves and with the A-UAVs to boost the content availability closer to the benchmark performance by approximately 9% (see Fig. 4b). It utilizes the currently visiting A-UAV's caching information and the preceding MF-UAV's caching decision to algorithmically select its own contents for caching, which is also shown in Fig. 3. Such selective caching will reduce the redundancy of multiple copies of the same content available through multiple sources at the same time. Difference in the effectiveness of selective caching can be observed in Fig. 4a and 4b, where caching decisions at MF-UAVs differ due to the difference in $R_{TAD}$ in both scenarios. Third, when the agent uses UCB exploration strategy, during the initial learning epochs the content availability increases promptly due to high upper confidence value of all contents, which avoids excessive exploitation. This is due to low sampling of requests. As learning progresses, the sparse request for unpopular contents keeps the upper confidence value high which maintains consistent exploratory behavior. Fig. 4a shows that such exploration strategy alone helps to boost the content availability closer to the benchmark performance by approximately 5% more than popular estimation-based methods [23]-[26].

Fourth, the adaptability of the proposed learning-based caching mechanisms is further emphasized in Fig. 4c. It showcases that the ability of *Top-k* MAB approach to learn the caching policy in a setting where the user preference changes over time. It goes on to highlight the reactive nature of the *Top-k* MAB + Selective Caching Policy, where content availability increases more promptly compared to the standalone *Top-k* MAB implementation. Note that dynamic user preference patterns are simulated using Smith-Waterman Distance-based sequence swapping [29] and changing the Zipf parameter (refer Equation 1).

Furthermore, Fig. 5 shows the convergence behavior of the *Top-k* MAB + Selective Caching in terms of access delay. It is observed that as learning progresses, the access delay for requested contents reduces while the content availability increases. This shows the improvement in learnt caching policy over the learning epochs and its effect on content access delay. It also shows the decrease in delay based on the reduction of *TAD* for certain range of contents. This is done to highlight the efficacy of the algorithm by showing the difference in cached contents and associated content availability for different *TAD*s. The best reduction in access delay is observed when Upper Confidence Bound (UCB) exploration is used at the *Top-k* MAB agent of A-UAVs and selective caching is applied at micro-ferrying UAVs.

## *B. Complexity and Performance Comparison*

To evaluate the computational efficiency and performance of the proposed *Top-k* Multi-Armed Bandit algorithm, we conducted a detailed comparison with six widely recognized methods for UAV-assisted content dissemination. They are Proactive Caching [17], Joint Optimization [25, 26], Named Data Networking (NDN) [16], Value-Based Caching (VBC) [29], Deep Reinforcement Learning (DRL) [26], and Graph-Theoretic Approaches [30] (both sparse and dense). Table II presents a comprehensive summary of these methods, that highlights computational complexity, and performance metrics, which includes content availability, adaptability, latency, and scalability.

Proactive Caching relies on static patterns or historical data to predict content demand. This approach achieves high content availability in stable environments where demand patterns remain predictable. Its computational complexity is $O(N^2 + M.k)$, where $N$ represents the total number of contents, and $M$ denotes user requests. Despite its efficiency in static systems, Proactive Caching faces limitations in adaptability and scalability, which makes it unsuitable for dynamic or highly variable environments.

In contrast, Joint Optimization methods target dynamic environments by simultaneously optimizing UAV trajectories and caching decisions. These methods achieve optimal cache placement by considering trajectory dynamics alongside content demand. However, this advantage comes at a significant computational cost of $O\left(N_{Trajectory}.C_A.N + T.N^2\right)$. The high complexity restricts scalability, posing challenges for real-time applications and large-scale systems. While Joint Optimization seeks to balance trajectory and caching dynamics, Named Data Networking (NDN) offers a more direct approach by matching user requests with cached content. The computational complexity of NDN is $O(M.\log k)$, that makes it both scalable and efficient in environments with stable request patterns. However, its reliance on predictable user behavior limits its adaptability in scenarios with heterogeneous or rapidly changing content demands.

To address scenarios with fluctuating demand, Value-Based Caching (VBC) prioritizes content using predefined metrics such as popularity and tolerable access delay. This approach balances content diversity and latency effectively. The computational complexity, $O(T.k + T_{recalc}.N)$, remains manageable for moderate-scale systems. However, the periodic recalculation of metrics introduces challenges in scalability and limits the method's adaptability in highly dynamic environments.

TABLE II.
COMPLEXITY AND PERFORMANCE COMPARISON

| Caching Strategy | Key Focus | Complexity | Relative Performance | Content Availability | Adaptability | Latency | Scalability |
|---|---|---|---|---|---|---|---|
| Proactive Caching | Predicts content demand based on historical data. | $O(N^2 + M.k)$ | ~ 87 % | High | Low | Low | Moderate |
| Joint Optimization | Simultaneous optimization of caching and trajectory. | $O\left(N_{Trajectory}.C_A.N + T.N^2\right)$ | ~ 87 % | High | Moderate | Moderate to High | Low |
| Named Data Networking (NDN) | Interest-based content caching. | $O(M.log\ k)$ | ~ 81 % | Moderate | Low | Moderate | High |
| Value-Based Caching (VBC) | Prioritizes content based on known popularity and delay. | $O(T.k\ +\ T_{recalc}.N)$ | ~ 91 % | High | Low | Moderate | Moderate |
| Deep Reinforcement Learning (DRL) | Learns caching policies dynamically using RL. | $O\left(T.N.d + T_{explore}.N\right)$ | ~ 89 % | High | High | Moderate to High | Low |
| Graph-Theoretic Approach (Sparse) | Models UAV network as a sparse graph. | $O\left(T.(N + N.\Delta)\right)$ | ~ 81 % | High | Moderate | Moderate | High |
| Graph-Theoretic Approach (Dense) | Models UAV network as a dense graph. | $O\left(T.(N^2 + N.\Delta)\right)$ | ~ 81 % | High | Low to Moderate | High | Low |
| *Top-k* Multi-Armed Bandit | Uses bandit learning to prioritize caching. | $O(T.N\log N)$ | ~ 89 % | High | High | Low | High |

Deep Reinforcement Learning (DRL) addresses adaptability by dynamically learning caching policies based on evolving content demands. Through reinforcement learning, DRL achieves high content availability and flexibility. However, the reliance on neural networks results in a computational complexity of $O\left(T.N.d + T_{explore}.N\right)$, where $d$ reflects the neural network complexity. DRL also requires precise state definitions, which depend on the designer's perspective [33, 34]. For UAV-assisted systems, potential state definitions could include content availability, proximity between UAVs, population density, or request rates. Defining effective states in real-world, dynamic environments remains a significant challenge, often reducing DRL's practical applicability.

Graph-Theoretic Approaches, unlike the other methods, focus on spatial coordination by modeling UAV networks as graphs. Nodes represent UAVs, contents, or user communities, while edges capture connectivity or mobility paths. Sparse graphs achieve a computational complexity of $O\big(T.(N + N.\Delta)\big)$, whereas dense graphs increase this to $O\big(T.(N^2 + N.\Delta)\big)$, where $\Delta$ denotes the average degree of nodes. While effective for spatial optimization, the computational cost rises significantly in dense networks, which makes these methods less suitable for dynamic and large-scale systems.

The proposed *Top-k* MAB algorithm avoids these challenges through a computational complexity of $O(T.N \log N)$, which efficiently ranks and selects the most suitable content for caching. This efficiency ensures scalability even in large systems. Unlike DRL or RL methods, *Top-k* MAB does not depend on state definitions, that simplifies its deployment in dynamic environments. Additionally, it operates without requiring prior knowledge of user preferences or content distributions, that enhances adaptability in real-time and heterogeneous conditions.

Compared to DRL, Joint Optimization, and Graph-Theoretic Approaches, the *Top-k* MAB algorithm offers comparable or superior performance at a fraction of the computational cost. Its logarithmic growth ensures scalability for large content pools, while the low complexity supports practical implementation in UAV-assisted systems designed for real-time operations.

This analysis demonstrates the unique balance achieved by the *Top-k* MAB algorithm between computational efficiency and high performance. Table II further illustrates these attributes, which provides a comparative view of the computational complexity and relative performance of all methods discussed in this study.

### *C. Convexity and Convergence Analysis*

The formulated optimization problem in this work involves maximizing content availability while minimizing access delay, constrained by low availability period of MF-UAVs. These objectives depend on non-linear and stochastic relationships derived from factors such as content popularity (modeled using a Zipf distribution), tolerable access delay and hover time. Consequently, the optimization problem is inherently non-convex, as observed in various scenarios where the solution space exhibits multiple local optima. To analyze the convexity of the problem, we leverage nonlinear programming (NLP) [35] and heuristic-based Genetic Algorithm (GA) [35] techniques to explore the solution space.

To verify the nature of the problem, we analyzed the optimization landscape under different parameter configurations. In scenarios where the problem is convex, as demonstrated in Fig. 6a, NLP effectively identifies global optima by leveraging surrogate optimization functions derived from [28], [29]. However, with changes in key parameters, the landscape becomes non-convex, as shown in Figs. 6b and 6c. These non-convex regions highlight the presence of multiple feasible solutions, that confirms the non-convexity of the problem in general. Here traditional approaches like NLP are insufficient. To explore and solve the problem in non-convex regions, heuristic-based GA is investigated. This method successfully identifies trade-off solutions within the Pareto-optimal front and validates its suitability for non-convex problems.

To ascertain the applicability of our proposed caching framework, the *Top-k* Multi-Armed Bandit algorithm integrates a hybrid exploration-exploitation strategy. The use of $\epsilon$-greedy exploration ensures the algorithm avoids getting trapped in local optima, while the UCB mechanism focuses on reward maximization. Furthermore, the multi-dimensional reward structure incorporates a holistic view of cache hits (see Section VI) and provides a robust approach to optimization in non-convex regions. The cumulative reward plot in Fig. 6d demonstrates the convergence of the *Top-k* MAB approach, with rewards stabilizing over time as the algorithm adapts to the dynamic solution space.

The optimization problem's nature is validated through the use of NLP for convex regions and GA for non-convex regions,

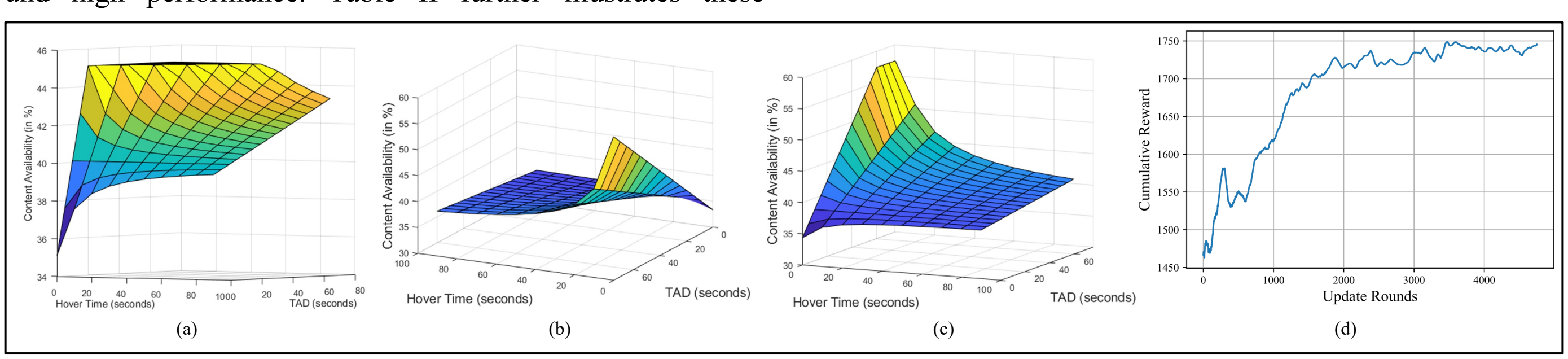

Fig. 6. (a) Convexity in objective space, (b)-(c) Non-convexity in objective space for parameters changes, (d) Convergence through reward

while the proposed *Top-k* MAB achieves robust convergence across both, which showcases its versatility and effectiveness.

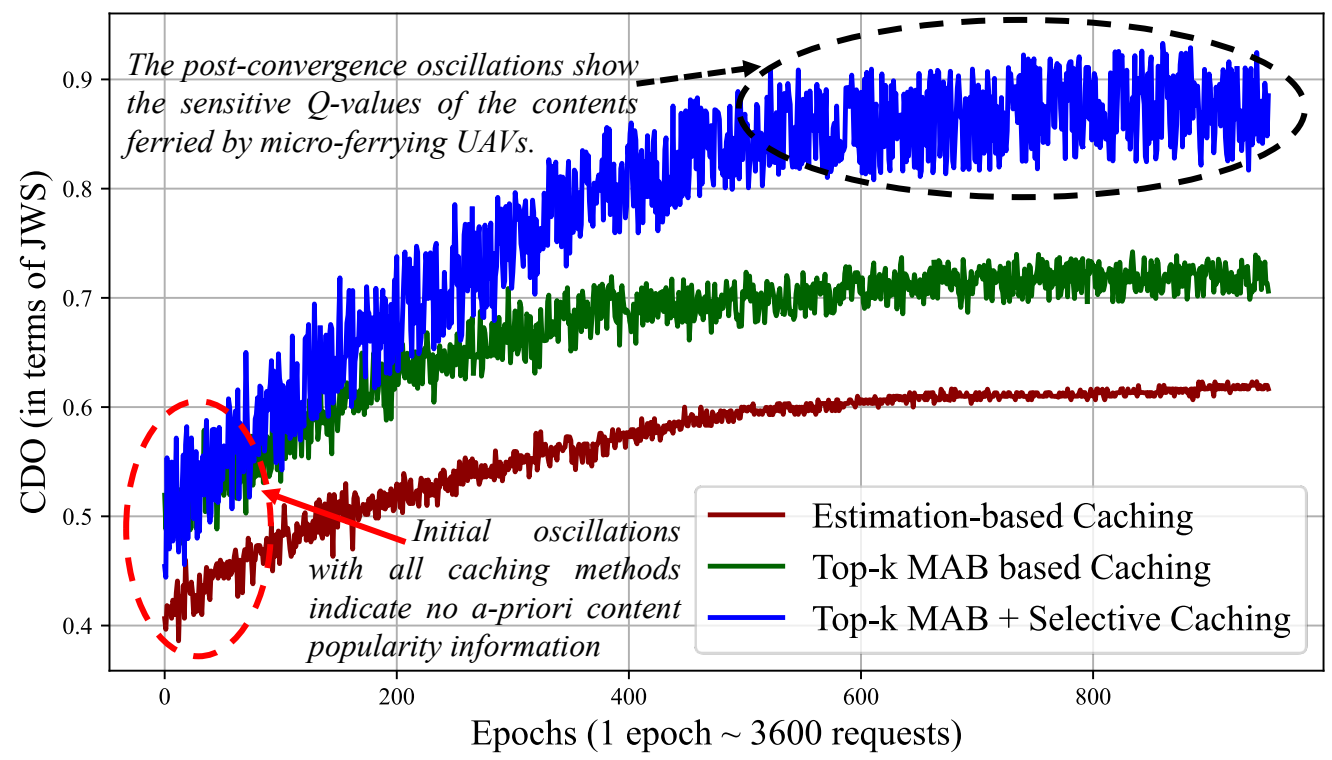

Fig. 7. Learnt cached content sequence's similarity with benchmark sequence

### D. Cache Similarity of Learnt Sequence with Best Sequence

The effects of learning on the cached content sequence are demonstrated in Fig. 7. It plots Cache Distribution Optimality (*CDO*) of the cached content sequences for all the A-UAVs in terms of Jaro-Winkler Similarity (*JWS*). The key observation are as follows. First, the average $CDO$ between the benchmark caching sequence from cache pre-loading policy (see Section V) and the cached content sequences learnt by the *Top-k* MAB agents at A-UAVs converge near 0.9, although with a certain variance. Physically, this represents higher degree of similarity after convergence, where 1 indicates complete similarity and 0 implies no similarity. Second, the cached contents improve over epochs as learning progresses. Lower $CDO$ values after the initial epochs signify that the A-UAVs have no *a priori* local or global content popularity information. As the MAB agents learn, over epochs of generated content requests, the cached contents in the A-UAVs become more similar to the best caching sequence. Third, $CDO$ is an indirect representation of the storage segmentation factor ($\lambda$), which is used to decide the segment sizes according to cache pre-loading policies [29]. A higher $CDO$ implies that, along with learning, the caching policy, the *Top-k* MAB agents learn to emulate the said segmentation behavior. Finally, the partial dissimilarity of the cached content sequence can be ascribed to the uncertainty (or regret) associated with the Q-values of contents with low popularity. Also, this leads to an oscillatory convergence of $CDO$ for the A-UAVs.

The impacts of selective caching at micro-ferrying UAVs can be distinctly seen in Fig 7. Selective caching at the MF-UAVs along with *Top-k* MAB caching agent at A-UAVs leads to a $CDO$ of nearly 0.9. Note that this depends on effective caching capacity of the MF-UAVs, which is dictated by the $TAD$s associated with content requests and the MF-UAVs visiting frequency at A-UAVs (refer Algorithm 2). The dependance of contents' Q-values on such information also adds to the post-convergence oscillation. To be noted that for the computation of $CDO$, the benchmark caching sequence is derived by considering the same effective caching capacity as the selective caching algorithm at the micro-ferrying UAVs.

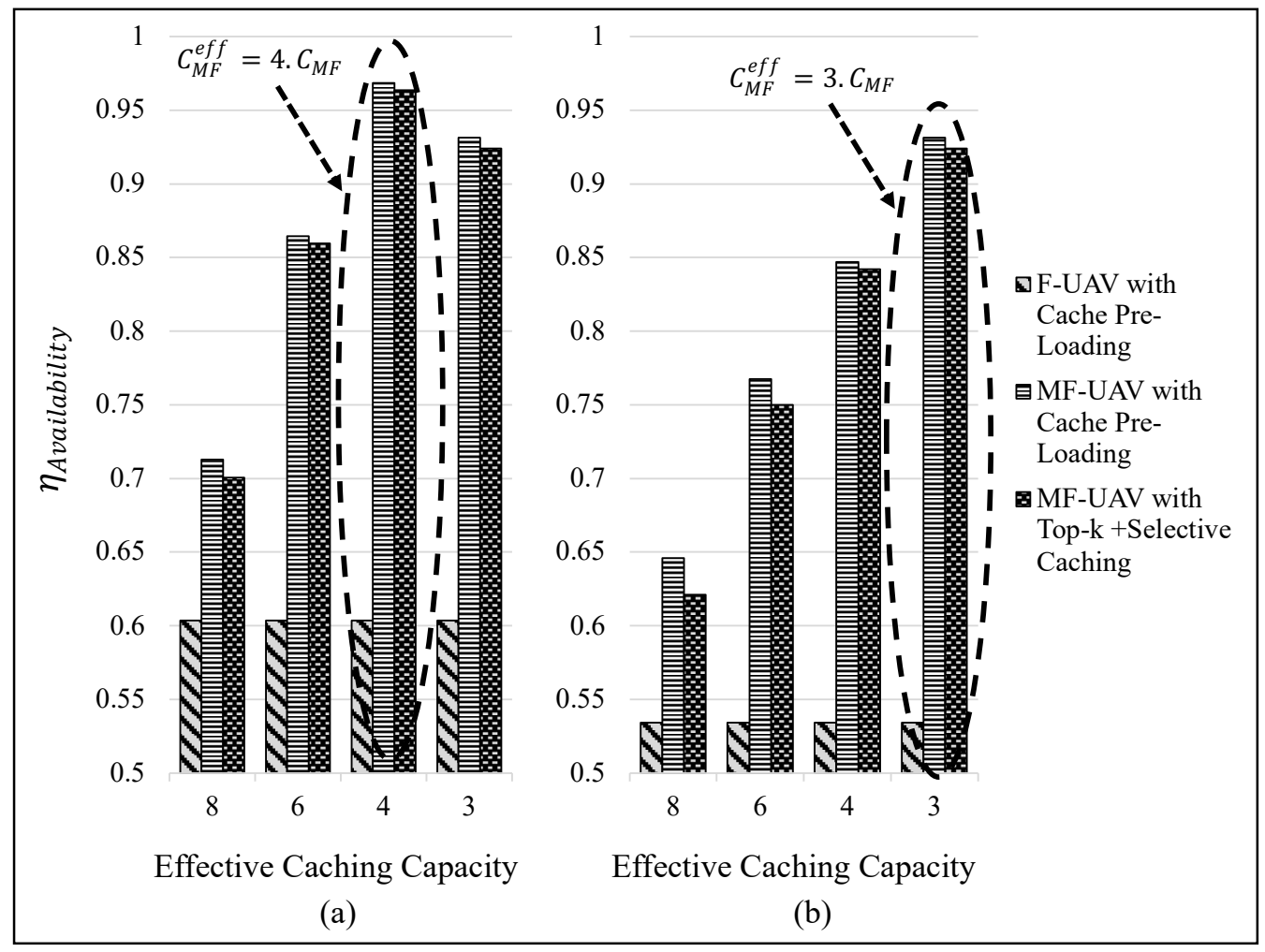

Fig. 8. (a) Best learnt $C_{MF}^{eff}$ for $R_{TAD} = 1/6$, (b) for $R_{TAD} = 1/8$

### E. Leveraging the Micro-Ferrying UAVs for Better Effective Caching Capacity

To elaborate on the ability of selective caching at micro-ferrying UAVs to exploit effective caching capacity, experiments are conducted with different *TAD* Ratios $R_{TAD}$. The comparison of performance is done with a scenario where there is one relatively larger ferrying UAV (F-UAV). Such F-UAVs can have sophisticated communication equipment as payload including a larger caching capacity (≥ total caching capacity of all MF-UAVs). The content availability according to the learnt caching policy with 24 MF-UAVs is shown in Fig. 8. The remaining parameters are set according to the default values provided in Table I. Following observations can be made from Fig. 8a. First, for a given $R_{TAD}$ =1/6, the best content availability achieved is with effective caching capacity of $4.C_{MF}$ i.e., four times the caching capacity of an MF-UAV. Physically, this means that the 4 MF-UAVs fly very close to each other. Within the fleet of such closely flying MF-UAVs none of the pending content requests, for the ones cached at the MF-UAVs, expire by exceeding their respective *TADs*. Second, content availability increases with increase in effective caching capacity up to a certain point beyond which it decreases with further increase in effective caching capacity. This is due to two opposing effects: a) low availability period [27] for a content increases with increase in effective caching capacity which eventually decreases content availability, and b) with increase in effective caching capacity content availability increases due to more types of contents cached at MF-UAVs. Therefore, selective caching at the MF-UAVs handles the trade-off between these opposing behaviors by choosing a caching policy that increases the effective caching capacity without increasing the low availability period of contents cached at MF-UAVs. Third, Fig. 8 shows how the caching policy adapts when MF-UAV groups change due to some MF-UAVs depleting their energy. The *Top-k* MAB, combined with Selective Caching, effectively adjusts to the uncertainties of MF-UAV downtime.

Note that the previous explanation is valid for a particular $R_{TAD}$. The best learnt effective caching capacity differs when the $TADs$ associated with the content requests change. This is demonstrated in Fig. 8b where due to a decrease in $R_{TAD}$ from 1/6 to 1/8, the best learnt effective caching capacity decreases. Therefore, it can be said that the learning capability of the *Top-k* MAB agents at A-UAVs have an indirect dependence on the effective caching capacity of the MF-UAVs.

This also emphasizes the motivation behind employing micro-UAVs in the role of ferrying contents. With a given cost budget for UAVs in a content dissemination system, micro-UAVs provide flexibility in caching policies such that their effective caching capacity can be altered to fit to the users' needs. This facility cannot be leveraged with relatively larger and pricier UAVs, especially under equipment cost constraints.

## VIII. Summary and Conclusion

In this paper, a micro-UAV aided content dissemination system is proposed which can learn caching policies on-the-fly without *a priori* content popularity information. Two types of UAVs are introduced for content provisioning in a disaster/war-stricken scenario viz. anchor UAVs and micro-ferrying UAVs. Cache-enabled anchor UAVs are stationed at each stranded community of users for uninterrupted content provisioning. Micro-ferrying UAVs act as content transfer agents across the anchor UAVs. A decentralized *Top-k* Multi-Armed Bandit Learning-based caching policy is proposed to ameliorate the limitation of existing caching methods. It learns the caching policy on-the-fly by maximizing the estimated multi-dimensional reward for the increase in local and global content availability. It is shown that a *Top-k* MAB learning based caching policy achieves a content availability of ≈82% of maximum achievable content availability. To improve the Q-value estimates, Selective Caching Algorithm is introduced at micro-ferrying UAVs. This method combines the shared information between anchor UAVs and micro-ferrying UAVs to reduce redundant copies of contents and to produce a better estimate of top popular content at a community. Selective caching at micro-ferrying UAVs along with *Top-k* MAB learning-based caching policy at anchor UAVs boosts the content availability to ≈89% of maximum achievable content availability. With the proposed caching policies, a scaled-up micro-UAV aided network is shown to attain a content availability of nearly 95% of maximum achievable content availability. The complexity analysis highlights the efficiency and practicality of the proposed *Top-k* MAB algorithm in dynamic UAV networks. Compared to existing methods, it provides a scalable and computationally efficient solution with high content availability, adaptability, and low latency. These characteristics make it ideal for real-time UAV-assisted content dissemination systems. Future work on this research includes algorithmically coping with time-varying content popularity and adaptive trajectory planning in the presence of operational unreliabilities of the UAV.

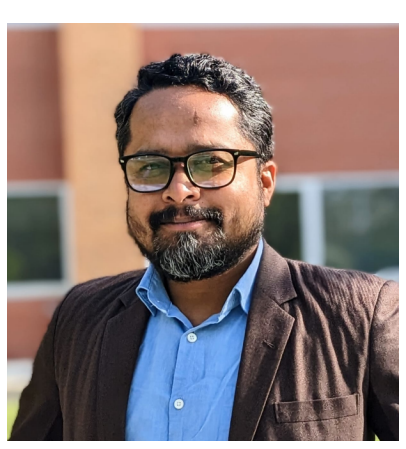

**Amit Kumar Bhuyan** (Graduate Student Member, IEEE) is currently working toward his PhD in ECE department at Michigan State University. His research interests include Distributed Systems, Vehicular Communication, Federated Learning, Reinforcement Learning, Recommender Systems and Signal Processing. He received the All-India Council of Technical Education Scholarship in 2014 and 2015.

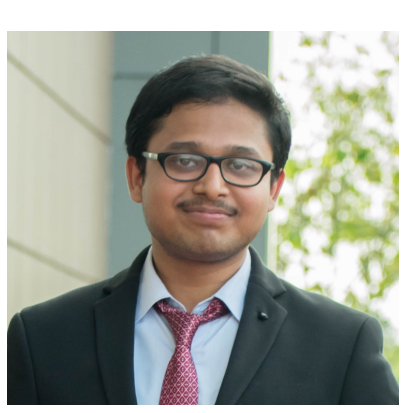

**Hrishikesh Dutta** (Graduate Student Member, IEEE) is a PhD student in the ECE department at Michigan State University. His research focuses on Distributed Systems, Medium Access Control and Reinforcement Learning. He is a recipient of S. N. Bose Scholars' Award 2017.

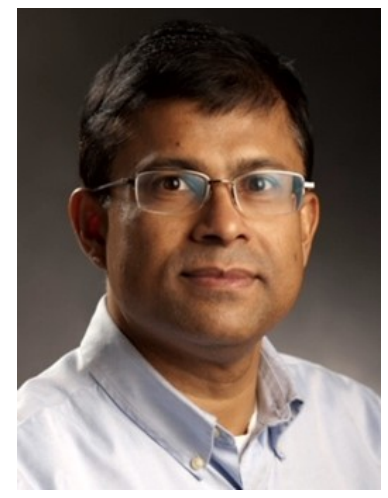

**Subir Biswas** is a professor (Senior Member, IEEE), and the director of the Networked Embedded and Wireless Systems laboratory at Michigan State University. He received his Ph.D. from University of Cambridge and held various research and management positions in NEC Research Institute, Princeton, AT&T Laboratories, Cambridge, and Tellium Optical Systems, NJ. His current research includes Data Dissemination in Vehicular Networks, Wearable Systems for Health Applications, and Learning in Embedded and Resource-constrained Systems.